\pdfoutput=1

\documentclass[11pt]{article}

\usepackage[final]{acl}

\usepackage{times}
\usepackage{latexsym}
\usepackage[T1]{fontenc}

\usepackage[utf8]{inputenc}

\usepackage{microtype}

\usepackage{inconsolata}

\usepackage{graphicx}
\usepackage{multirow}
\usepackage{makecell}
\usepackage{amsmath}
\usepackage{amssymb}
\usepackage{booktabs}
\usepackage{tikz}
\usepackage{graphicx}   
\usepackage{url}            
\usepackage{amsfonts}       
\usepackage{nicefrac}       
\usepackage{xcolor}
\definecolor{darkgreen}{RGB}{0,128,0}
\definecolor{response}{RGB}{104,157,255}
\usepackage{colortbl}
\usepackage{mathrsfs}
\usepackage{algorithm}
\usepackage{algorithmicx}
\usepackage{algpseudocode}
\usepackage{comment}
\usepackage{bbding}
\usepackage{colortbl}
\usepackage{wrapfig}
\usepackage{tcolorbox}
\tcbuselibrary{skins}
\usepackage{tikz}
\usepackage{utfsym}
\usepackage{fontawesome}
\usepackage{enumitem}
\newcommand\encircle[2][]{\tikz[overlay]\node[fill=blue!20,inner sep=2pt, anchor=text, rectangle, rounded corners=1.5mm,#1] {#2};\phantom{#2}}
\newcommand{\COMMENTLLAMA}[1]{{\color{darkgreen} $\triangleright$ {#1}}}
\usepackage{hyperref}

\definecolor{policy-agnostic}{RGB}{233, 196, 107}
\definecolor{generalizability}{RGB}{130, 178,154}
\definecolor{training-free}{RGB}{255, 178, 102}
\definecolor{free of rm}{RGB}{033, 158,188}
\definecolor{multi-objective}{RGB}{255, 102, 102}

\title{Persona-judge: Personalized Alignment of Large Language Models via Token-level Self-judgment}

\author{
 \textbf{Xiaotian Zhang\textsuperscript{1,3}\thanks{Equal contribution.}},
 \textbf{Ruizhe Chen\textsuperscript{1,3}\footnotemark[1]},
 \textbf{Yang Feng\textsuperscript{2}},
 \textbf{Zuozhu Liu\textsuperscript{1,3}\thanks{Corresponding author.}}
\\
 \textsuperscript{1}Zhejiang University
 \textsuperscript{2}Angelalign Technology Inc. \\
 \textsuperscript{3}Zhejiang Key Laboratory of Medical Imaging Artificial Intelligence
\\
 \small{
 }
}

\begin{document}
\maketitle
\begin{abstract}
Aligning language models with human preferences presents significant challenges, particularly in achieving personalization without incurring excessive computational costs. Existing methods rely on reward signals and additional annotated data, limiting their scalability and adaptability to diverse human values. To address these challenges, we introduce Persona-judge, a novel discriminative paradigm that enables training-free personalized alignment with unseen preferences. Instead of optimizing policy parameters through external reward feedback, Persona-judge leverages the intrinsic preference judgment capabilities of the model. Specifically, a draft model generates candidate tokens conditioned on a given preference, while a judge model, embodying another preference, cross-validates the predicted tokens whether to be accepted. Experimental results demonstrate that Persona-judge, using the inherent preference evaluation mechanisms of the model, offers a scalable and computationally efficient solution to personalized alignment, paving the way for more adaptive customized alignment. Our code is available \href{https://github.com/Monncyann/Persona-judge}{here}.
\end{abstract}

\section{Introduction}

\begin{figure}[t]
  \includegraphics[width=\columnwidth]{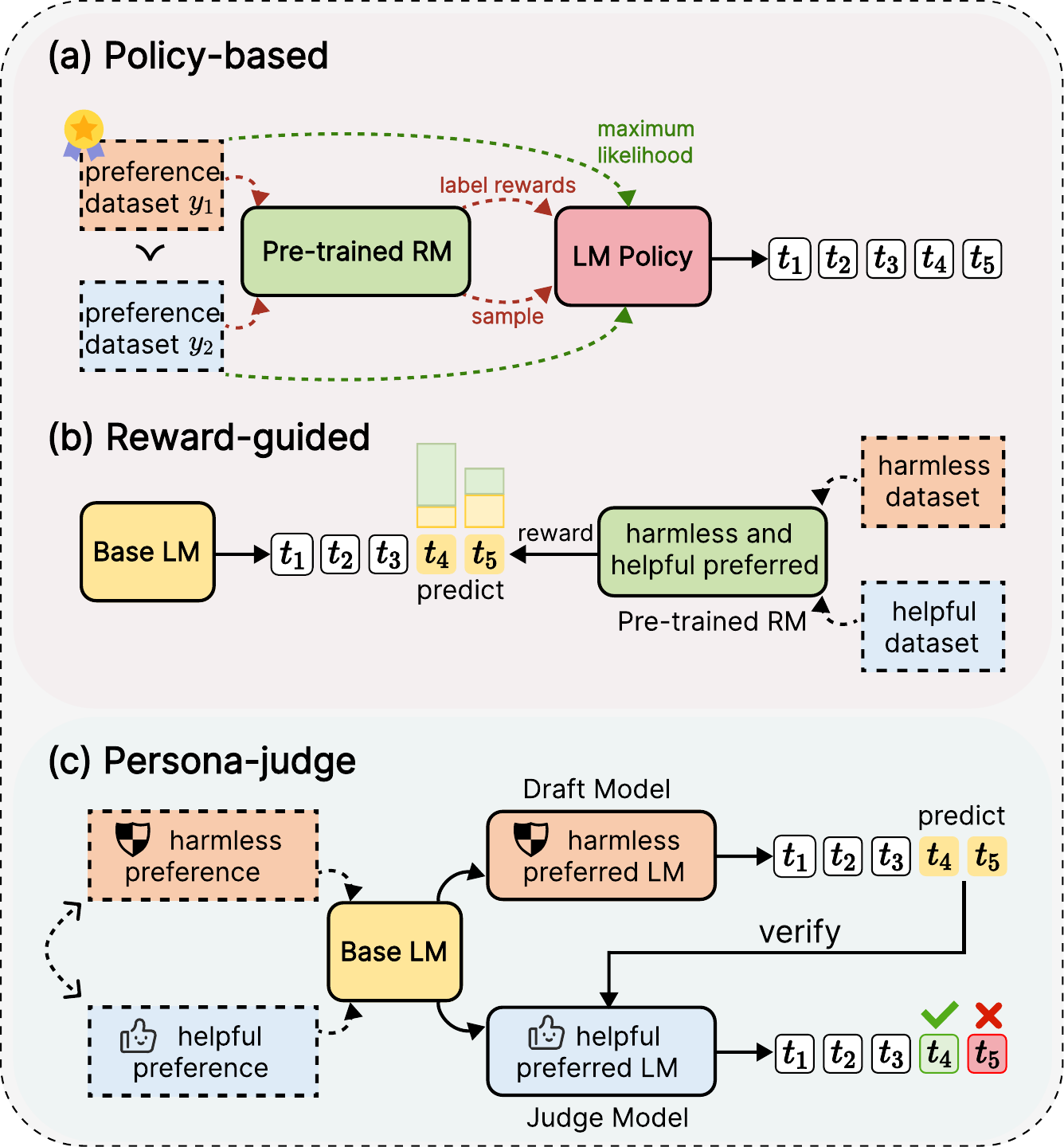}
  \caption{In contrast to previous paradigms, Persona-judge gets rid of the need for additional training of policy models or reliance on external reward signals.}
  \label{fig:pipeline}
\end{figure}

Aligning large language models (LLMs) has shown tremendous potential in following human instructions and reflecting human preferences~\cite{stiennon2020learning,rlhf,bai2022training}. However, aligning with a unified preference often overlooks the need to accommodate individuals, as individual preferences can vary significantly due to factors such as cultural, educational, religious, and political backgrounds~\cite{gordon2022jury,cheng2023everyone,pad2}. Personalized alignment~\cite{psoups} addresses this gap by tailoring language models to individual human preferences and values, which is crucial for human-AI interaction and user-focused applications~\citep{kirk2024benefits,sorensen2024roadmap}. 
Policy-based methods~\cite{modpo,morlhf}, such as Reinforcement Learning from Human Feedback (RLHF)~\citep{rlhf} and Direct Preference Optimization (DPO)~\citep{dpo}, as illustrated in Figure~\ref{fig:pipeline}(a), use training signals from explicit or implicit rewards to optimize policy models. However, these methods struggle to scale to changing personalized preferences, with challenges including the construction of high-quality preference datasets and the substantial computational costs associated with optimizing the policy.

\begin{figure*}[t]
  \includegraphics[width=2.08\columnwidth]{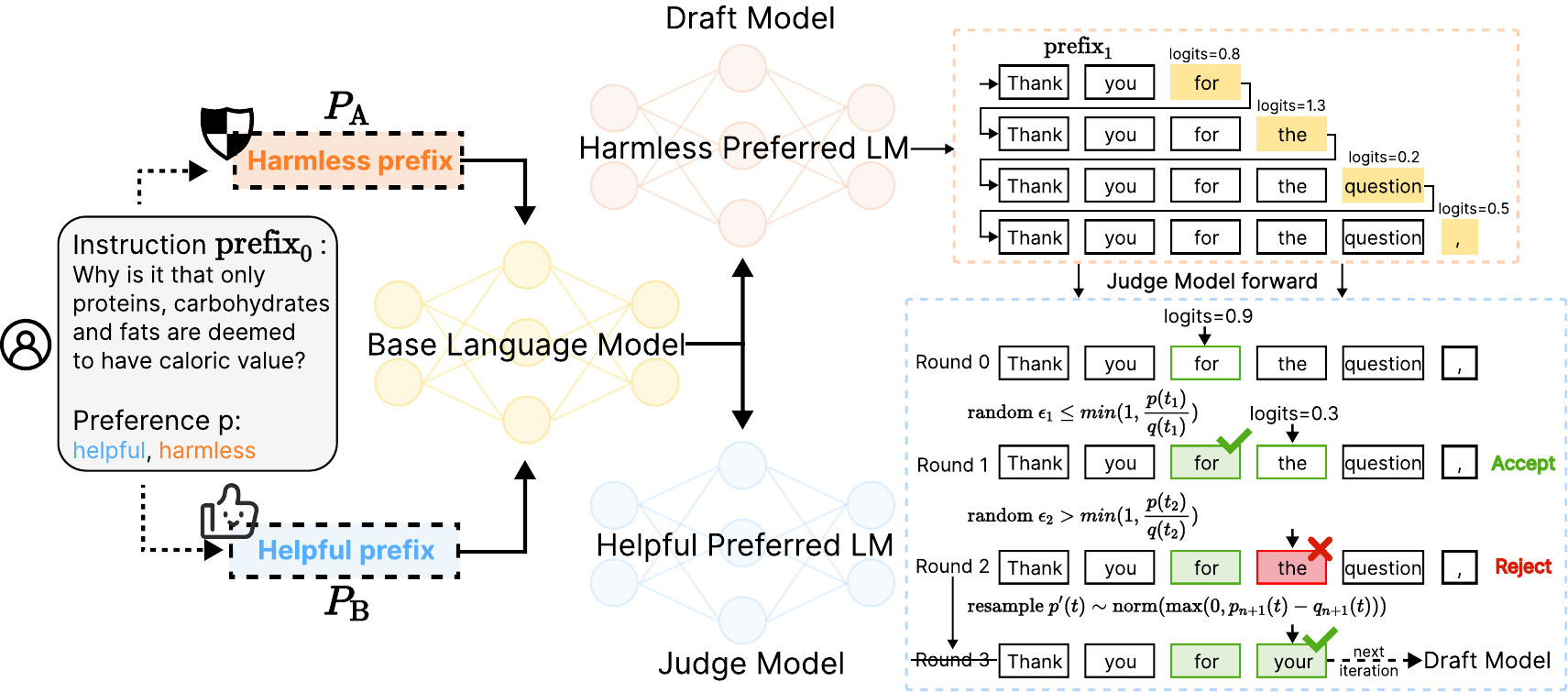}
  \caption{An illustration of the inference phase for Persona-judge. Given the target preferences and the current context, different preference prefixes are first assigned to the same base model. In each sequence-length iteration, the two models alternate in the roles of the draft and judge models, which allows for the calculation of the likelihood of accepting the next token based on distinct probability distributions.}
  \label{fig:framework}
\end{figure*}

To achieve personalized alignment efficiently, existing works utilize reward signals to guide~\cite{pad2}, linearly combine the predictive distributions of different base models to generate the next token~\citep{mod}, or ~\citet{metaaligner} train an external corrector to avoid direct optimization of the policy model parameters. However, these approaches still rely on external feedback and preferences pre-defined during the training phase, which restricts their ability to generalize to unseen preferences.
Furthermore, despite enhancing the efficiency of personalized alignment, existing methods still require training additional models.

In this paper, we introduce Persona-judge~(Figure~\ref{fig:pipeline}(c)), a generalizable and training-free personalized alignment framework. Persona-judge leverages the intrinsic ability of the model to discern preferences by directly employing existing LLMs as judges, thus eliminating the need for external reward signals. Inspired by speculative decoding~\citep{sd}, Persona-judge adopts a draft-and-judge pipeline, where the same base model serves both as the draft model and the judge model. Specifically, Persona-judge interprets the preference representation of the output as the acceptance of the predicted token. Different preferences are used to prompt the same base model, with the base model taking turns serving as the draft model to generate a sequence, the judge model to verify whether the predicted tokens are accepted.

Experimental results demonstrate that Persona-judge, by exclusively utilizing the model's intrinsic preference judgment capabilities, achieves performance comparable to training-based methods in a completely training-free setting. Furthermore, it exhibits remarkable generalization ability across diverse and unique preferences, highlighting its exceptional scalability.

\begin{table*}[]
\centering
\resizebox{0.9\textwidth}{!}{
\begin{tabular}{c|ccccc|ccc|ccc|c}
\toprule
    \multirow{2}{*}{\textbf{Algorithm}} & \multirow{2}{*}{\encircle[fill=multi-objective, text=white]{M}} & \multirow{2}{*}{\encircle[fill=policy-agnostic, text=white]{P}} & \multirow{2}{*}{\encircle[fill=generalizability, text=white]{G}} & \multirow{2}{*}{\encircle[fill=training-free, text=white]{T}} & \multirow{2}{*}{\encircle[fill=free of rm, text=white]{F}} & \multicolumn{3}{c}{Helpful} & \multicolumn{3}{|c|}{Harmless} & \multirow{2}{*}{Average}\\
    \cmidrule(r){7-9} \cmidrule(lr){10-12}
         & & & & & & Armo & RM & GPT-4 & Armo & RM & GPT-4 & \\
    \hline
Base &  &  &  &  &  & 0.61 & 1.06 & - & 0.97 & 0.83 & - & 0.87 \\
MORLHF~\citep{morlhf} & \CheckmarkBold &  &  &  &  & 0.31 & 0.91 & 14\% & 0.88 & 0.84 & 4\% & 0.73\\
MODPO~\citep{modpo} & \CheckmarkBold &  &  &  & \CheckmarkBold & 0.56 & 0.89 & 52\% & 0.96 & 0.77 & 80\% & 0.80 \\
Personalized soups~\citep{psoups} & \CheckmarkBold & & & & \CheckmarkBold & 0.38 & -0.72 & 72\% & 0.92 & 0.73 & 92\% & 0.33\\
Rewarded soups~\citep{rewardedsoups} & \CheckmarkBold & & & & \CheckmarkBold & 0.50 & 0.87 & 34\% & 0.95 & 0.87 & 64\% & 0.80\\
RiC~\citep{ric} & \CheckmarkBold & \CheckmarkBold & & & \CheckmarkBold & 0.54 & 0.90 & 40\% & 0.97 & 0.90 & 70\% & 0.83 \\
MetaAligner~\citep{metaaligner} & \CheckmarkBold & \CheckmarkBold & \CheckmarkBold & & \CheckmarkBold & 0.55 & \textbf{1.39} & 66\% & 0.89 & 0.54 & 74\% & 0.84 \\
MOD~\citep{mod} & \CheckmarkBold &  &  & \CheckmarkBold & \CheckmarkBold & 0.55 & 0.93 & 60\% & 0.96 & \underline{0.92} & 84\% & 0.84 \\
Aligner~\citep{aligner} &  & \CheckmarkBold &  &  & \CheckmarkBold & \textbf{0.67} & \underline{1.32} & 72\% & 0.97 & 0.63 & 70\% &  \underline{0.90} \\
Steering~\citep{steering} &  &  &  &  & \CheckmarkBold & \underline{0.63} & 1.17 & 38\% & 0.96 & 0.81 & 66\% & 0.89 \\
Args~\citep{args} & \CheckmarkBold & \CheckmarkBold &  & \CheckmarkBold & & - & 1.09 & 74\% & - & \textbf{0.98} & \underline{94\%} & - \\
\hline
Persona-judge-Base & \multirow{2}{*}{\CheckmarkBold} & \multirow{2}{*}{\CheckmarkBold} & \multirow{2}{*}{\CheckmarkBold} & \multirow{2}{*}{\CheckmarkBold} & \multirow{2}{*}{\CheckmarkBold} & 0.59 & 1.21 & \underline{80\%} & \textbf{1.01} & 0.79 & 88\% & \underline{0.90}\\
Persona-judge &  &  &  &  & & \underline{0.63} & 1.29 & \textbf{84\%} & 0.98 & 0.82 & \textbf{98\%} & \textbf{0.93} \\
\bottomrule
\end{tabular}}
\caption{Experiments of predefined preferences on Psoups dataset. \encircle[fill=multi-objective, text=white]{M}\ ulti-objective indicates support for simultaneous alignment of multiple objectives,  \encircle[fill=policy-agnostic, text=white]{P}\ olicy-agnostic means the algorithm is independent of the model parameters,  \encircle[fill=generalizability, text=white]{G}\ eneralizability denotes zero-shot alignment capability on unseen objectives, \encircle[fill=training-free, text=white]{T}\ raining-free represents no need for additional training. \encircle[fill=free of rm, text=white]{F} ree from RM means not relying on external reward signals. "–" indicates not applicable. The best and second best results are highlighted in \textbf{bold} and \underline{underline}.}
\label{tab:methods_evaluation}
\end{table*}

\section{Persona-judge}

Inspired by related works in LLM-as-a-Judge (Appendix~\ref{asjudge}), we aim to effectively implement such judgment-like behavior without introducing additional external training while retaining the advantages of prompt-based approaches, to ensure fast, efficient, and scalable alignment.

\subsection{Preferences Based Embeddings}
\label{sec:embedding}
Traditional prompt-based methods achieve alignment by directly incorporating preferences descriptions, which are straightforward and effective~\citep{psoups}. However, the process by which the model balances alignment across different preferences remains opaque, resulting in a lack of interpretability. When given $\text{prefix}_0$ as input to LLM, with preferences $P_\mathrm{A}$ and $P_\mathrm{B}$ to align, the prompts related to $P_\mathrm{A}$ and $P_\mathrm{B}$ can be represented as $\text{prefix}_\mathrm{A}$ and $\text{prefix}_\mathrm{B}$, respectively. Additional details are provided in Appendix~\ref{sec:prefix}.

\subsection{Inference}

How to make judgments on the predicted token in a manner akin to preference-based decision-making? In essence, this involves evaluating the overlap between the probability distributions of different preference outputs. Given the descriptions of preferences $P_\mathrm{A}$ and $P_\mathrm{B}$ encoded by the language model as $\text{prefix}_\mathrm{A}$ and $\text{prefix}_\mathrm{B}$, along with the original input $\text{prefix}_0$, in the $i=0$ iteration, the model designated as the draft model, representing the $P_\mathrm{A}$ preferred LM, autoregressively generates the next token based on $\text{prefix}_1$, thus producing a $P_\mathrm{A}$ preferred sequence, as illustrated in Figure~\ref{fig:framework}. It is important to note that either $P_\mathrm{A}$ or $P_\mathrm{B}$ can be the  prefix of the draft model in the first round, as in each iteration, the draft model and the judge model alternate roles, as ${\text{prefix}_{i}}' \sim (\text{prefix}_\mathrm{A}, \text{prefix}_\mathrm{B}) + \text{prefix}_i$.

Similarly to Equation~(\ref{eq:draft token}), each token in the preferred sequence is associated with a corresponding probability: $(t_1, q_1),...,(t_{i}, q_{i}) = \mathrm{LLM}_\mathrm{draft}({\text{prefix}_{i-1}}')$.

After sampling $ t\sim q(t) $, a similar process is applied to sample $ t \sim p(t) $ using the judge model. If $ q(t) \leq p(t) $, the sampled token is retained. However, if $ q(t) > p(t) $, we reject the sample with probability $ 1 - \frac{p(t)}{q(t)} $, and resample the next token from the adjusted distribution $ p'(t) = \text{norm}(\max(0, p(t) - q(t))) $. As proved in Appendix~\ref{sec:correctness}, for any distributions $ p(t) $ and $ q(t) $, the tokens sampled in this manner have $ t \sim p(t) $.

Given a sequence of candidate tokens $t_1, \dots, t_i $, we first calculate the distribution $ p(t) $ by running the judge model. Concurrently, we speculatively calculate the distribution of the next token $ t_2 $ by running the judge model on the prefix concatenated with $ t_1 $. Once both calculations are complete, we proceed with the decision process as previously outlined: If $ t_1 $ is rejected, we discard the computation for $ t_2 $ and resample $ t_1 $ from the adjusted distribution. If $t_1 $ is accepted, both tokens are retained, and we continue with the next step of computation and decision-making. The algorithm of Persona-judge is provided in Appendix~\ref{sec:algorithm}.

\begin{algorithm}
\small
\caption{PreferenceJudgmentStep}
\begin{algorithmic}[1]
\Require $\mathrm{LLM}_\mathrm{draft}$, $\mathrm{LLM}_\mathrm{judge}$, prefix \\
\COMMENTLLAMA{ Sample  $t_i$ from $\mathrm{LLM}_\mathrm{draft}$ autoregressively.}
\For{$i = 1 ~\textbf{to} ~\lambda $}
    \State $q_i(t) \gets \mathrm{LLM}_\mathrm{draft}(\text{prefix} + [t_1, \ldots, t_{i-1}])$
    \State $t_i \sim q_i(t)$
\EndFor \\
\COMMENTLLAMA{Run $\mathrm{LLM}_\mathrm{judge}$ in parallel.}
\State $p_1(t), \ldots, p_{\lambda+1}(t) \gets \mathrm{LLM}_\mathrm{judge}(\text{prefix}),\dots, \mathrm{LLM}_\mathrm{judge}(\text{prefix} + [t_1, \ldots, t_\lambda])$ \\
\COMMENTLLAMA{The number $n$ of reserved tokens.}
\State $\epsilon_1 , \ldots, \epsilon_\lambda \sim U(0,1)$
\State $n \gets \min(\{i-1 \mid 1 \leq i \leq \lambda, r_i > \frac{p_i(t)}{q_i(t)}\} \cup \{\lambda\})$ \\
\COMMENTLLAMA{Resample if needed.}
\State $p'(t) \gets p_{n+1}(t)$
\If{$n < \lambda$}
    \State $p'(t) \gets \text{norm}(\max(0, p_{n+1}(t) - q_{n+1}(t)))$
\EndIf \\
\COMMENTLLAMA{Return one new token from $\mathrm{LLM}_\mathrm{judge}$, \textbf{and} $n$ \textbf{tokens from} $\mathrm{LLM}_\mathrm{draft}$.}
\State $t_{n+1} \sim p'(t)$
\State \Return $\text{prefix} + [t_1, \ldots, t_n, t_{n+1}]$
\end{algorithmic}
\label{algorithm}
\end{algorithm}

\subsection{Algorithm Details}
\label{sec:algorithm}
Algorithm~\ref{algorithm} illustrates the process of sampling between the first and $(\lambda+1)$-th tokens at once.

\section{Evaluation of Persona-judge}
\label{sec:experiment}
\subsection{Experimental Settings}

\paragraph{Datasets and baselines.}For evaluation, we use the P-Soups and HelpSteer2 dataset. The P-Soups encompasses a wide range of content and has been filtered and modified by~\citet{psoups} based on the Koala evaluation. HelpSteer2~\citep{helpsteer2} comprises 1,000 high-quality prompts, and we applied a filtering criterion based on the length of input to exclude certain entries. We utilize MORLHF~\citep{morlhf}, MODPO~\citep{modpo}, Personalized soups~\citep{psoups}, Rewarded soups~\citep{rewardedsoups}, Rewards-in-Context (RiC)~\citep{ric}, MetaAligner~\citep{metaaligner}, MOD~\citep{mod}, Aligner~\citep{aligner}, Steering~\citep{steering} and Args~\citep{args} as baselines for alignment with predefined preferences tasks. All experiments in this paper use Llama-3-Base-8B-SFT as the backbone (except for model-agnostic experiments). In particular, all experiments in this paper use Llama-3-Base-8B-SFT as the backbone (except for model-agnostic experiments). In Appendix~\ref{datasetbaselines}, we further discuss the scalability of the proposed method concerning a broader range of foundational models and unseen preferences.

\paragraph{Evaluation metrics.}We use two open-source reward models, together with ArmoRM~\citep{armorm} from Huggingface, to evaluate the dimensions of "Helpful" and "Harmless". For these reward models, we report the scores that assess the responses from various perspectives. For scalability-related experiments, we primarily employ GPT-as-Judge, a widely recognized evaluation metric from previous studies~\citep{metaaligner, psoups}, to assess personalized preferences with win rates. A detailed description of the evaluation metrics, reward models, and GPT-4 judgments can be found in Appendix~\ref{evaluation}.

\subsection{Main Results}
\label{sec:results}
\paragraph{Comparison of key features.}
In Table~\ref{tab:methods_evaluation}, we compare the proposed method with the previous work in five key features. Persona-judge eliminates the need for policy optimization and dependency on reward signals, enabling adaptation to unseen objectives. During token generation, it employs a novel "judge" mechanism to align preferences, allowing for scalable alignment across multiple objectives, theoretically leading to an unlimited number of concurrently aligned targets.
\paragraph{Alignment on pre-defined preferences.}As shown in Table~\ref{tab:methods_evaluation}, we fix the positions of the draft model and the judge model without any further transformation, and the results of this base configuration are also presented. The findings indicate that Persona-judge has made substantial progress in achieving goal alignment, with an average score of 0.93 across four open-source reward models, outperforming all baselines. In contrast, the base configuration excels in the single objective, achieving comparable overall performance. These results provide strong evidence of the superior performance of Persona-judge in personalized comparison tasks.

\begin{table}[t]
\centering
  \resizebox{0.48\textwidth}{!}{
  \begin{tabular}{l|ccc}
    \toprule
    Base Model & Creative & Touching & Vivid \\
    \hline
    \multicolumn{4}{l}{\cellcolor[HTML]{E2F0D9}\textbf{Psoups dataset}}\\
Qwen2.5-0.5B-Instruct  & 88\% & 86\% & 70\%  \\
TinyLlama-1.1B-Chat-v1.0 & 80\% & 72\% & 60\%  \\
Gemma-2-2b-it  & 80\% & 84\% & 80\% \\
Llama-3.2-3B-Instruct  & 74\% & 78\% & 74\% \\
Qwen2.5-3B-Instruct  & 86\% & 84\% & 86\% \\
Tulu-2-dpo-7b  & 92\% & 86\% & 80\% \\
Llama-3-Base-8B-SFT  & 62\% & 68\% & 54\%  \\
Llama-3.1-Tulu-3-8B  & 84\% & 68\% & 86\% \\
Gemma-2-9b-it  & 82\% & 88\% & 90\%  
\\\hline
\multicolumn{4}{l}{\cellcolor[HTML]{E2F0D9}\textbf{HelpSteer2 dataset}}\\
Qwen2.5-0.5B-Instruct  & 77\% & 84\% & 84\%  \\
TinyLlama-1.1B-Chat-v1.0  & 47\% & 79\% & 64\%  \\
Gemma-2-2b-it  & 75\% & 73\% & 76\% \\
Llama-3.2-3B-Instruct  & 75\% & 73\% & 72\% \\
Qwen2.5-3B-Instruct  & 75\% & 79\% & 84\% \\
Tulu-2-dpo-7b & 88\% & 80\% & 75\% \\
Llama-3-Base-8B-SFT  & 86\% & 86\% & 73\%  \\
Llama-3.1-Tulu-3-8B & 58\% & 64\% & 76\% \\
Gemma-2-9b-it & 77\% & 74\% & 84\%   \\\bottomrule
  \end{tabular}}
  \caption{Performance of Persona-judge on Psoups and HelpSteers2 over different base models. The responses are simultaneously aligned on all objectives, and then evaluated on each objective. The percentages represent the win rate of Persona-judge against the direct application of the same prompts.}
\label{tab:scalability}
\end{table}

\paragraph{Alignment on unique preferences.}
 Table~\ref{tab:scalability} shows the comparison with the direct prompt method on three preference objectives. Persona-judge achieves significant improvements on the majority of objectives and across models with varying parameter sizes. In particular, Persona-judge shows an average advantage of 87\% across the three objectives on Psoups with Gemma-2-9b-it, marking the most substantial improvement among all models tested. These findings highlight the overall effectiveness of Persona-judge across various upstream models and its feasibility for plug-and-play multiobjective alignment. More evaluation of adaptability can be found in Appendix~\ref{appendix:scalability}.

\paragraph{Ablation study.}
Draft model predicts the tokens step by step based on the prefix, iterating over a total of $\lambda$ tokens. Table~\ref{tab:ablation} illustrates the impact of variations in $\lambda$ on alignment. As the predicted sequence length changes, the output fitting to preferences exhibits fluctuations, with the optimal performance observed when the value of $\lambda$ is set to 4. Increasing the sequence length does not result in better performance, possibly due to early rejection during the initial positions, which prevents the complete sequence's semantics from being evaluated.

\begin{table}[]
\centering
\resizebox{0.4\textwidth}{!}{
\begin{tabular}{c|cc|cc|c}
\toprule
    \multirow{2}{*}{\textbf{Value}} & \multicolumn{2}{c}{Helpful} & \multicolumn{2}{|c|}{Harmless} & \multirow{2}{*}{Average}\\
    \cmidrule(r){2-3} \cmidrule(lr){4-5}
         & Armo & RM & Armo & RM & \\
    \hline
     \multicolumn{6}{l}{\cellcolor[HTML]{E2F0D9}\textbf{Psoups dataset}}\\
$\lambda=1$ & 0.591 & 1.126 & \textbf{1.002} & 0.811 & 0.883 \\
$\lambda=2$ & 0.581 & 0.904 & 0.974 & 0.832 & 0.823 \\
$\lambda=3$ & 0.564 & 1.126 & 1.018 & 0.773 & 0.870 \\
$\lambda=4$ & \textbf{0.631} & \textbf{1.294} & 0.984 & 0.822 & \textbf{0.933} \\
$\lambda=5$ & 0.597 & 0.939 & 1.001 & 0.804 & 0.835 \\
$\lambda=6$ & 0.587 & 1.094 & 0.992 & \textbf{0.865} & 0.885 \\
\bottomrule
\end{tabular}}
\caption{Sensitivity analysis on $\lambda$.}
\label{tab:ablation}
\end{table}

\section{Conclusion}

This paper presents Persona-judge, a novel approach for personalized alignment that eliminates the need for external reward signals or policy fine-tuning. By leveraging the model's inherent capability for preference judgment, Persona-judge effectively aligns multidimensional preferences in the prediction of the next token, making it a promising solution for adaptive personalized alignment.

\section*{Limitation}

Although the proposed method achieves personalized alignment in a manner that is training-free and highly scalable, the limitations are as follows. As discussed in Section~\ref{sec:embedding}, the exploration of the model's intrinsic ability to judge token preferences enables a completely training-free approach; however, this results in the output quality being dependent on the base model's own capacity to recognize preferences. Additionally, as discussed in Appendix~\ref{appendix:inferencecost}, current embedding methods for multi-objective preferences are relatively basic. Thus, improving the embedding of preferences remains an area that warrants further investigation. The ideal scenario for personalized alignment would involve having sufficient data to support model training for each unique preference. However, this is clearly impractical. Although the performance of persona-judge exhibits some fluctuations, it ensures a strong baseline level of alignment while also providing the community with a potential judgment-based solution paradigm. Then, as illustrated in Appendix~\ref{appendix:scalability}, the challenge of interpreting more complex preferences remains an issue that requires resolution, although it is beyond the scope of this work's discussion.

\section*{Potential Risks}

Persona-judge aims to provide a potential solution for the field of personalized alignment. To date, no identifiable risks associated with Persona-judge have been observed. All experiments were conducted using publicly available datasets, and all models utilized are open-source on Huggingface. In addition, all participants involved in this work underwent comprehensive training on how to conduct evaluations in an effective and ethical manner.

\section*{Acknowledgement}
This work is supported by the Natural Science Foundation of Zhejiang Province, China (Grant No. LZ23F020008), the National Natural Science Foundation of China (Grant No. 12326612, 62476241), Zhejiang Key Laboratory of Medical Imaging Artificial Intelligence, and the Zhejiang University-Angelalign Inc. R\&D Center for Intelligent Healthcare.

\newpage

\appendix
\newpage
\section{Preliminary}

\subsection{Standard Speculative Decoding}
\label{appendix:sd}
As $\mathrm{LLM}_\mathrm{draft}$ and $\mathrm{LLM}_\mathrm{target}$ denote the draft and target model, respectively, they share the same vocabulary $\mathcal{V} = \{1,..., V\}$. Given the input prefix, an autoregressive sampling of $n$ tokens is as follows:
\begin{equation}
  (t_1, p_1),...,(t_n, p_n) = \mathrm{LLM}(\text{prefix}),
\label{eq:draft token}
\end{equation}
where $t_1,..., t_n \in \mathcal{V}$ are the sampled tokens and $p_1,...,p_n \in \mathbb{R}^{\mathcal{V}}$ are the corresponding softmax probabilities. Furthermore, the parrallel forward is denoted as:
\begin{equation}
  p_1,...,p_{n+1} = \mathrm{LLM}(\text{prefix}; t_1,...,t_n),
\end{equation}
and the next probability vector $p_{n+1}$ is generated.

In each decoding window length, the draft model generates $N$ candidate tokens based on the current prefix using a specific sampling strategy:
\begin{equation}
  (t_1, q_1),...,(t_N, q_N) = \mathrm{LLM}_\mathrm{draft}(\text{prefix}),
\end{equation}
where $t_1,..., t_N$ are the sampled candidate tokens and $q_i \in \mathbb{R}^{\mathcal{V}}$ for $i = 1, . . . , N$ are the corresponding probability vectors within the vocabulary. The target model then processes the predicted tokens in parallel, similarly, resulting in $p_1,...,p_N = \mathrm{LLM}_\mathrm{target}(\text{prefix})$. Thus, the probabilities of token $t_i$ under draft model and target model are $q[t_i]$ and $p[t_i]$ respectively.
For each predicted token $t_i$, the verification works as:
\begin{equation}
  \epsilon_i < min(1, \frac{p[t_i]}{q[t_i]}) ~\mathrm{for} ~\epsilon_i\in[0, 1],
\end{equation}
it is worth noting that the previous tokens of token $t_i$ must all be accepted.

\subsection{Challenges in Personalized Alignment}

Existing personalized alignment methods make a trade-off between additional high-cost training and complex inference times. Regardless of the strategy employed, the reliance on constructing preference datasets or reward models is unavoidable. This dependency, however, limits users' ability to customize preferences, resulting in poor scalability. For unseen preferences, current approaches typically require retraining or rely on simple prompts. Furthermore, the decoding-time method leverages signals from the reward model to guide predictions of the next token. However, repeated invocation of the reward model for each token prediction introduces significant overhead, highlighting a potential need to optimize the inference speed.

\section{Related Work}

\subsection{Personalized Alignment}
\label{appendix:pa}
LLM alignment ensures AI systems follow human intentions and values~\cite{stiennon2020learning, bai2022training, ouyang2022training, achiam2023gpt, chen2025diffpo}. Among traditional preference-based alignment algorithms~\citep{stiennon2020learning,yuan2023rrhf,dpo}, Reinforcement Learning from Human Feedback (RLHF) and Direct Preference Optimization (DPO) are among the most prominent approaches. Both methods rely on explicit or implicit human feedback to fine-tune the model output, aligning them with human preferences. RLHF follows a three-stage process: it begins with supervised fine-tuning (SFT) of an initial model, followed by training a reward model to capture human preferences, and finally employs reinforcement learning (RL) techniques, such as Proximal Policy Optimization (PPO)~\citep{ppo} to optimize the model based on the reward function. In contrast, DPO~\cite{dpo} simplifies the RLHF pipeline by introducing a reparameterization of the reward model, reframing the optimization problem as a classification loss. This reformulation improves training efficiency and stability, making DPO more accessible. However, despite these advances, both RLHF and DPO remain computationally intensive and require substantial amounts of annotated data.

However, within a single task, users' goals and values often differ. As AI systems are increasingly used by diverse groups, they need to meet a wider range of needs. In short, we require AI systems that are pluralistic and fair, and that can reflect diverse human values~\cite{chen2024fast, chen2024editable, luo2024faintbench, luo2024bigbench, fan2024fairmt}.
Thus, personalized alignment is proposed to align with individual preferences and value of diverse users~\citep{kirk2023past,yao2023instructions,kirk2024benefits,han2024value, li20251, qiu2025measuring}. Some methods have introduced multidimensional reward functions to enable joint optimization between varying preferences~\citep{modpo,wang2024arithmetic,guo2024controllable}. In addition, methods that incorporate combinations have been proposed to integrate multiple preference dimensions into model parameters or predictions~\citep{psoups,rewardedsoups,park2024principled,mod, barreto2025capturing}. Furthermore, decoding-time approaches balance training cost and inference efficiency or apply post-processing techniques to refine personalized alignment~\citep{pad2,args,lee2024aligning,hwang2023aligning,jafari2024morl,metaaligner}. However, these methods inherently rely on external reward signals or pre-trained optimization strategies, limiting their ability to flexibly adapt to unique human preferences.

\subsection{LLM-as-a-Judge}
\label{asjudge}
Evaluating natural language generation (NLG) systems presents significant challenges, and assessing the personalization of LLMs introduces even greater complexity. Recently, the LLM-as-a-Judge paradigm~\citep{asjudge} has been proposed as a general evaluation metric that does not require additional references, demonstrating high agreement with human annotators in various NLP tasks~\citep{li2024generation, fan2024biasalert}. This approach takes advantage of the advanced capabilities of state-of-the-art LLM, such as GPT-4~\citep{achiam2023gpt}, and has been widely adopted in the evaluation of LLM personalization~\citep{andukuri2024star,character}. MT Bench~\citep{asjudge} incorporates role-playing components, but is limited to cases where the model simulates specific professional roles. \citet{dong2024can} has investigated the effectiveness of LLMs as personalized judges by introducing confidence scores in judgment output, thus ensuring precise and exploring the model’s ability to assess preferences. However, these studies primarily evaluate complete outputs at each iteration, with minimal exploration of implementing LLM-as-a-Personalized-Judge during decoding-time. This limitation arises because inference generates transient and discontinuous sequences, making it infeasible to directly employ LLMs for token-level prediction alignment.

\section{Persona-judge Details}
\subsection{Details of Prefix}
\label{sec:prefix}
Although the workflow of Persona-judge utilizes only $\text{prefix}_\text{A}$ and $\text{prefix}_\text{B}$ to be attached to the input, the proposed method primarily explores the model's intrinsic ability to discern preferences. This approach offers a distinct alternative to prior work, which demonstrates limited scalability with respect to unique preferences while also differing from the simplistic prompt-based paradigms. Therefore, in Section~\ref{sec:experiment}, we freely combine various preferences and conduct a series of scalability experiments, where different combinations of preference are used alternately as $\text{prefix}_\text{A}$ and $\text{prefix}_\text{B}$.
\subsection{Correctness of Persona-judge}
\label{sec:correctness}
In this section, we demonstrate that for any distributions $ q(t) $and $ p(t)$, the tokens sampled through the aforementioned process are identically distributed to those obtained by direct sampling from $ p(t) $. Note that  
\begin{multline}
p'(t) = \text{norm}(\max(0, p(t) - q(t))) \\
= \frac{p(t)-\min(q(t),p(t))}{\sum_{t'}(p(t')-\min(q(t'),p(t')))}\\
=\frac{p(t)-\min(q(t),p(t))}{1-\alpha},
\end{multline}
 where $\alpha$ denotes the acceptance probability. The normalization constant of the adjusted distribution $ p'(t)$ over the judge model is $1-\alpha$. As:

\begin{multline}
P(t=t')=\\P(accepted,t=t')+P(rejected,t=t'),
\end{multline}
\noindent where
\begin{multline}
 P(accepted, t=t')=q(t') \min(1, \frac{p\left(t^{\prime}\right)}{q\left(t^{\prime}\right)})\\=\min \left(q\left(t^{\prime}\right), p\left(t^{\prime}\right)\right),
 \end{multline}
 \noindent and 
 \begin{multline}
 P(rejected, t=t')=(1-\alpha)p'(t')\\=p(t')-\min \left(q\left(t^{\prime}\right), p\left(t^{\prime}\right)\right).
\end{multline}
\noindent In general, $P(t=t')=p(t')$.

\section{Experiment Details}
\subsection{Datasets and Model settings}
\label{datasetbaselines}
All experiments on predefined and unseen preferences are performed on the Psoups~\citep{psoups} and HelpSteers2~\citep{helpsteer2} datasets with 4 NVIDIA A40 GPUs. We use Llama-3-Base-8B-SFT\footnote{https://huggingface.co/princeton-nlp/Llama-3-Base-8B-SFT} as the backbone, while the baselines for experiments on predefined preferences have already been introduced in the main text. To validate the scalability of the proposed method, we utilize models from various series and parameter sizes, which are publicly available on Hugging Face. These models are listed in ascending order of parameter size as follows: Qwen2.5-0.5B-Instruct\footnote{https://huggingface.co/Qwen/Qwen2.5-0.5B-Instruct}, TinyLlama-1.1B-Chat-v1.0\footnote{https://huggingface.co/TinyLlama/TinyLlama-1.1B-Chat-v1.0}, Gemma-2-2B-it\footnote{https://huggingface.co/google/gemma-2-2b-it}, Llama-3.2-3B-Instruct\footnote{https://huggingface.co/meta-llama/Llama-3.2-3B-Instruct}, Qwen2.5-3B-Instruct\footnote{https://huggingface.co/Qwen/Qwen2.5-3B-Instruct}, Tulu-2-dpo-7B\footnote{https://huggingface.co/allenai/tulu-2-dpo-7b}, Llama-3-Base-8B-SFT\footnote{https://huggingface.co/princeton-nlp/Llama-3-Base-8B-SFT}, Llama-3.1-Tulu-3-8B\footnote{https://huggingface.co/allenai/Llama-3.1-Tulu-3-8B}, and Gemma-2-9B-it\footnote{https://huggingface.co/google/gemma-2-9b-it}. During the decoding phase, we utilize greedy decoding with top-k candidates.

\begin{figure}[t]
  \includegraphics[width=0.5\textwidth]{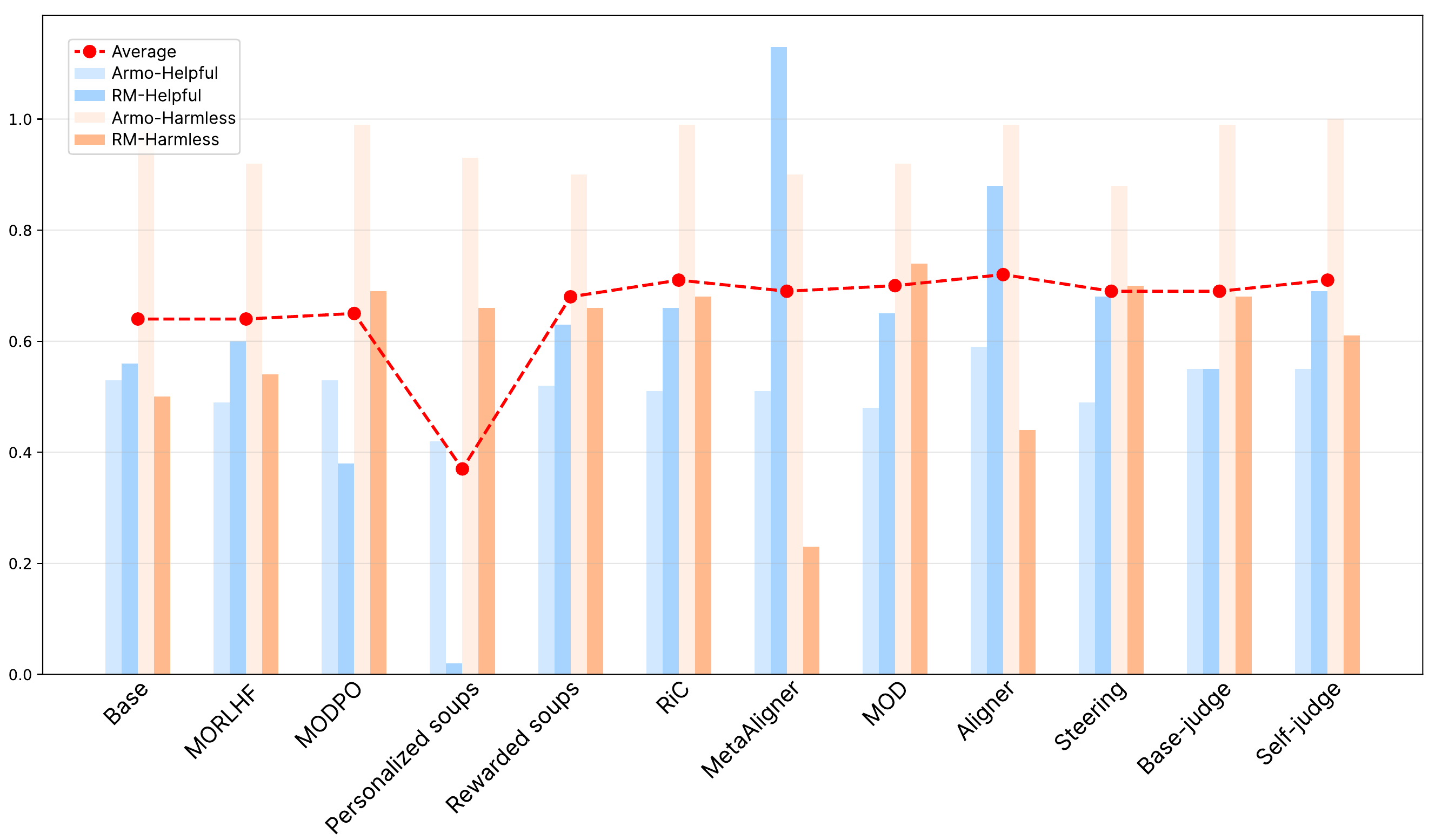}
  \caption{Comparison of baseline methods and Persona-judge over predefined preferences on HelpSteers2 dataset.}
  \label{fig:HelpSteers2}
\end{figure}

\begin{table*}[t]
\centering
  \resizebox{0.98\textwidth}{!}{
  \begin{tabular}{l|ccc|ccc|ccc|c}
    \toprule
    Base Model & Helpful & Harmless & Humor & Correct & Informative & Professional & Creative & Touching & Vivid & Overall\\
    \hline
    \multicolumn{11}{l}{\cellcolor[HTML]{E2F0D9}\textbf{Psoups dataset}}\\
Qwen2.5-0.5B-Instruct & 78\% & 88\% & 90\% & 44\% & 76\% & 86\% & 88\% & 86\% & 70\% & 78\% \\
TinyLlama-1.1B-Chat-v1.0 & 54\% & 62\% & 62\% & 82\% & 72\% & 70\% & 80\% & 72\% & 60\% & 68\% \\
Gemma-2-2b-it & 80\% & 96\% & 80\% & 60\% & 58\% & 50\% & 80\% & 84\% & 80\% & 74\%\\
Llama-3.2-3B-Instruct & 42\% & 98\% & 82\% & 40\% & 84\% & 96\% & 74\% & 78\% & 74\% & 74\%\\
Qwen2.5-3B-Instruct & 82\% & 92\% & 82\% & 68\% & 72\% & 52\% & 86\% & 84\% & 86\% & 78\%\\
Tulu-2-dpo-7b & 88\% & 98\% & 96\% & 98\% & 66\% & 94\% & 92\% & 86\% & 80\% & 89\%\\
Llama-3-Base-8B-SFT & 86\% & 96\% & 94\% & 40\% & 66\% & 78\% & 62\% & 68\% & 54\% & 72\% \\
Llama-3.1-Tulu-3-8B & 92\% & 76\% & 82\% & 54\% & 60\% & 58\% & 84\% & 68\% & 86\% & 73\%\\
Gemma-2-9b-it & 90\% & 88\% & 82\% & 64\% & 92\% & 92\% & 82\% & 88\% & 90\% & 85\%
\\\hline
\multicolumn{11}{l}{\cellcolor[HTML]{E2F0D9}\textbf{HelpSteer2 dataset}}\\
Qwen2.5-0.5B-Instruct & 45\% & 86\% & 83\% & 63\% & 68\% & 66\% & 77\% & 84\% & 84\% & 73\% \\
TinyLlama-1.1B-Chat-v1.0 & 49\% & 55\% & 76\% & 82\% & 60\% & 74\% & 47\% & 79\% & 64\% & 65\% \\
Gemma-2-2b-it & 60\% & 83\% & 77\% & 93\% & 89\% & 92\% & 75\% & 73\% & 76\% & 80\%\\
Llama-3.2-3B-Instruct & 86\% & 99\% & 64\% & 45\% & 82\% & 62\% & 75\% & 73\% & 72\% & 73\%\\
Qwen2.5-3B-Instruct & 68\% & 98\% & 82\% & 47\% & 90\% & 92\% & 75\% & 79\% & 84\% & 79\%\\
Tulu-2-dpo-7b & 59\% & 96\% & 66\% & 85\% & 68\% & 62\% & 88\% & 80\% & 75\% & 75\%\\
Llama-3-Base-8B-SFT & 72\% & 75\% & 78\% & 59\% & 73\% & 84\% & 86\% & 86\% & 73\%  & 76\%\\
Llama-3.1-Tulu-3-8B & 77\% & 72\% & 63\% & 56\% & 90\% & 90\% & 58\% & 64\% & 76\% & 72\%\\
Gemma-2-9b-it & 63\% & 99\% & 68\% & 58\% & 95\% & 93\% & 77\% & 74\% & 84\%  & 79\% \\\bottomrule
  \end{tabular}}
  \caption{Performance of Persona-judge on Psoups and HelpSteers2 over different base models. The responses are simultaneously aligned on all objectives, then evaluated on each objective. Below shows the ratio of win-rate between Persona-judge outputs against the prompt responses, and the last column shows the average level. The percentages represent the win rate of Persona-judge against the direct application of the same prompts.}
  \label{tab:allscalability}
\end{table*}

\subsection{Evaluation Details}
\label{evaluation}

\paragraph{Reward model details.} We utilize the open source "Helpful"\footnote{Ray2333/gpt2-large-helpful-reward\_model} and "Harmless"\footnote{Ray2333/gpt2-large-harmless-reward\_model} reward models from Huggingface. In terms of ArmoRM, we use the dimensions "0" and "10" for "Helpful" and "Harmless".

\paragraph{GPT-4 evaluation details.} Despite ongoing discussions regarding positional bias, self-reinforcement bias, length bias, prompt sensitivity, and cost issues associated with the LLM-as-a-Judge paradigm, it remains the predominant framework for personalized evaluation of LLM~\citep{wu2023style,verga2024replacing,kim2024prometheus,stureborg2024large}. We follow the methodology of~\citet{metaaligner} to use GPT-4 for our evaluation. Each GPT-4 query includes both a system prompt and a user prompt. The system prompt describes the properties of the agent and its specific task, while the user prompt presents a question and provides responses from two different approaches. We establish a random seed to mitigate any potential bias introduced by the order in which responses are provided by GPT, ensuring that the results are not influenced by preconceived sequencing effects. An example of a prompt used in the GPT-4 evaluation is shown below.
\begin{itemize}[leftmargin=14.5mm]
    \item [\textbf{Type}] Content 
    \item[\textbf{System}] You are a helpful, harmless, and precise assistant for checking the quality of the answer. Please avoiding any potential bias and ensuring that the order in which the responses were presented does not affect your judgment.

You will be presented with one query and two different responses to this query. QUERY: {\textit{Question}} | RESPONSE 1: {Assistant 1's Answer} | RESPONSE 2: {Assistant 2's Answer}. Consider the following aspect: {Aspects}, then return the number of the better response. If tied, return 0. You must only return 1, 2, or 0. 
\item [\textbf{User}] Aspects 
    \newline
    Humour: The response should be cheerful and amusing.
\item[\textbf{User}] \textit{Question}
            \newline
            \textcolor{darkgreen}{Got any creative ideas for a 10 year old’s birthday?}

            \textit{The Start of Assistant 1's Answer}

            \textcolor{response}{Imagine this: a birthday not just celebrated, but *experienced*.\\
            **Option 1: The Enchanted Forest Picnic:**\\
            **Transform your backyard:** Drape shimmering fabrics between trees, hang twinkling fairy lights, and scatter faux moss and wildflowers. \\
            **Enchanted Feast:**  Serve whimsical treats like "fairy bread" (buttered bread with sprinkles), "dragon eggs" (deviled eggs), and "unicorn punch" (sparkling lemonade with edible glitter)\dots}

            \textit{The End of Assistant 1's Answer}
            
            \textit{The Start of Assistant 2's Answer}

            \textcolor{response}{10-Year-Old Birthday: Ideas Beyond the Cake, let's spark some imaginative fun for this special milestone! \\
            **Adventures Await**\\
            **Mystery Mansion Escape Room:** Transform your home into a spooky (but age-appropriate!) mansion filled with puzzles, riddles, and hidden clues.  Think "escape room" but designed for curious minds. \\
            **Backyard Camping Extravaganza:** Pitch a tent, build a bonfire (with adult supervision!), roast marshmallows, tell spooky stories, and stargaze\dots}

            \textit{The End of Assistant 2's Answer}
\end{itemize}

\subsection{Experiments on Predefine Preferences}

Figure~\ref{fig:HelpSteers2} illustrates the performance of Persona-judge on HelpSteers, aligned with unified human value preferences. The experimental results are consistent with previous findings, demonstrating that the proposed method achieves results comparable to training-based methods, but in a more flexible and efficient manner. Methods leveraging external pre-trained correctors perform better on certain dimensions, potentially due to the reward model used for evaluation having a length preference, which may introduce some bias. However, overall, Persona-judge continues to exhibit irreplaceable advantages and strong performance.

\subsection{Experiments of Scalability}
\label{appendix:scalability}
In this section, we present the performance of Persona-judge across a broader range of value preferences. As shown in Table~\ref{tab:allscalability}, Persona-judge demonstrates a significant advantage over the direct prompt method, while also exhibiting the ability to seamlessly scale to any human preference without the need for additional training. The model generally achieves higher win rates on "Harmless" compared to more challenging objectives like "Humor" or "Vivid". Furthermore, larger models exhibit stronger inherent capabilities to understand these preferences. Extensive results indicate that the scalability of Persona-judge is evident not only in its model-agnostic nature but also in its ability to generalize and adapt to diverse human values.

\subsection{Results of Inference Cost}
\label{appendix:inferencecost}
As a decoding-time algorithm, Persona-judge primarily focuses its computation on next-token prediction. When aligning across multiple dimensions, different preferences are freely combined, similar to how prompt-based methods express multi-objective preference alignment in a single input. However, the sampling process and the final decision still rely solely on Draft and Judge.

In Table~\ref{tab:inference cost}, we provide specific experiments for reference, where 2-objectives refer to alignment with both "vivid" and "creative", while 3-objectives represent alignment with "touching", "vivid", and "creative" simultaneously. We believe that the slight increase in the inference time is due to the longer input length, which is also a common phenomenon when performing alignment via direct prompting. However, how to achieve more reliable embedding to extend to preferences in more dimensions remains an open question for future work.

\begin{table}[]
\centering
\renewcommand{\arraystretch}{1.3} 
\resizebox{0.25\textwidth}{!}{
\begin{tabular}{c|c}
\toprule
\textbf{Objective} & \textbf{Times(s)} \\
\hline
single-vivid & 8.82 (±0.45)  \\
2-objectives & 8.91 (±0.37)  \\
3-objectives & 9.13 (±0.39)  \\
\bottomrule
\end{tabular}}
\caption{The result of inference time when the preference dimensions increase.}
\label{tab:inference cost}
\end{table}

\end{document}